\documentclass[twoside]{article}

\usepackage[accepted]{aistats2020}
\usepackage[utf8]{inputenc} % allow utf-8 input
\usepackage[T1]{fontenc}    % use 8-bit T1 fonts
\usepackage{hyperref}       % hyperlinks
\usepackage{url}            % simple URL typesetting
\usepackage{booktabs}       % professional-quality tables
\usepackage{amsfonts}       % blackboard math symbols
\usepackage{nicefrac}       % compact symbols for 1/2, etc.
\usepackage{microtype}      % microtypography

\usepackage{enumerate}
\usepackage{natbib}
\usepackage{multirow}
\usepackage{makecell}
\usepackage{amsmath, amsthm, amssymb,graphicx}
\usepackage{xcolor}
\usepackage{appendix}

\usepackage{mathtools}
\usepackage{comment}
\usepackage{thmtools}
\usepackage{thm-restate}

\usepackage{xfrac}

\usepackage{subfigure}
\usepackage[utf8]{inputenc} % allow utf-8 input
\usepackage[T1]{fontenc}    % use 8-bit T1 fonts
\usepackage{url}            % simple URL typesetting
\usepackage{booktabs}       % professional-quality tables
\usepackage{amsfonts}       % blackboard math symbols
\usepackage{nicefrac}       % compact symbols for 1/2, etc.
\usepackage{microtype}      % microtypography
\usepackage{amsmath}
\usepackage{algorithm,algorithmic}
\usepackage{graphicx}
\usepackage{amsmath,amssymb,amsthm}
\usepackage{algorithm,algorithmic}

\usepackage{enumitem}
\setlist[description]{leftmargin=0.1in,labelindent=0in}

\usepackage{xcolor}
\usepackage{framed}
\usepackage{graphicx}
\graphicspath{ {images/} }
\usepackage{tikz}

\usepackage[T1]{fontenc}
\usepackage[utf8]{inputenc}
\usepackage{tabularx,ragged2e,booktabs,caption}
\newcolumntype{C}[1]{>{\Centering}m{#1}}

\usepackage{caption}

\usepackage{cleveref}

\usepackage{thmtools}
\usepackage{thm-restate}

\usepackage{graphicx}
\usepackage{wrapfig}

\newtheorem{theorem}{Theorem}

\usepackage[acronym,smallcaps,nowarn]{glossaries}
\glsdisablehyper{}
\newacronym{ES}{es}{evolution strategies}
\newacronym{RL}{rl}{reinforcement learning}
\newacronym{PG}{pg}{policy gradient}
\newacronym{MDP}{mdp}{{M}arkov decision process}
\newacronym{MC}{mc}{{M}onte {C}arlo}

\begin{document}



\twocolumn[

\aistatstitle{Variance Reduction for Evolutionary Strategies via Structured Control Variates}

\aistatsauthor{ Yunhao Tang \And Krzysztof Choromanski \And  Alp Kucukelbir }

\aistatsaddress{ Columbia University \And  Google Robotics \And Fero Labs \& Columbia University } ]

% ALP: do not use \gls in abstract as we need to copy-paste this into CMT
\begin{abstract}
\Gls{ES} are a powerful class of blackbox optimization techniques that recently became a competitive alternative to state-of-the-art \gls{PG} algorithms for \gls{RL}. We propose a new method for improving accuracy of the ES algorithms, that as opposed to recent approaches utilizing only Monte Carlo structure of the gradient estimator, takes advantage of the underlying \gls{MDP} structure to reduce the variance.
We observe that the gradient estimator of the ES objective can be alternatively computed using reparametrization and \gls{PG} estimators, which leads to new control variate techniques for gradient estimation in \gls{ES} optimization. We provide theoretical insights and show through extensive experiments that this \gls{RL}-specific variance reduction approach outperforms general purpose variance reduction methods. 
%Evolution strategies (ES) are a powerful class of blackbox optimization techniques for reinforcement learning (RL) that recently become a compelling alternative to state-of-the-art policy gradient (PG) methods due to its conceptual simplicity and structure admitted to highly distributed computational models. We propose a variance reduction strategy to improve the performance of ES. In contrast to standard Monte Carlo variance reduction techniques, we develop a control variate that takes advantage of the underlying Markov decision process in RL tasks. This provides a new connection between ES and PG methods. We study the theoretical properties of our control variate and show through extensive experiments that this RL-specific variance reduction approach outperforms general-purpose variance reduction methods over a variety of benchmarks.
\end{abstract}

\glsresetall{}
\section{Introduction}
\Gls{ES} have regained popularity through their successful application to modern \gls{RL}. \gls{ES} are a powerful alternative to \gls{PG} methods. Instead of leveraging the \gls{MDP} structure of a given \gls{RL} problem, \gls{ES} cast the \gls{RL} problem as a blackbox optimization. To carry out this optimization, \gls{ES} use gradient estimators based on randomized finite difference methods. This presents a trade-off: \gls{ES} are better at handling long term horizons and sparse rewards than \gls{PG} methods, but the \gls{ES} gradient estimator may exhibit prohibitively large variance.

Variance reduction techniques can make both methods more practical. Control variates (also known as baseline functions) that leverage Markovian \citep{mnih2016,schulman2015,schulman2017} and factorized policy structures \citep{gu2015muprop,liu2017action,grathwohl2017backpropagation,wu2018variance}, help to improve \gls{PG} methods. In contrast to these structured approaches, variance reduction for \gls{ES} has been focused on general-purpose \gls{MC} techniques, such as antithetic sampling \citep{salimans2017evolution,mania2018simple}, orthogonalization \citep{choromanski2018structured}, optimal couplings \citep{rowland2018geometrically} and quasi-\gls{MC} sampling \citep{choromanski2018structured,rowland2018geometrically}.

\textbf{Main idea.} We propose a variance reduction technique for \gls{ES} that leverages the underlying \gls{MDP} structure of \gls{RL} problems. We begin with a simple re-parameterization of the problem that uses \gls{PG} estimators computed via backpropagation. We follow by constructing a control variate using the difference of two gradient estimators of the same objective. The result is a \gls{RL}-specific variance reduction technique for \gls{ES} that achieves better performance across a wide variety of \gls{RL} problems.

\begin{figure*}[h]
\centering
\includegraphics[width=4in]{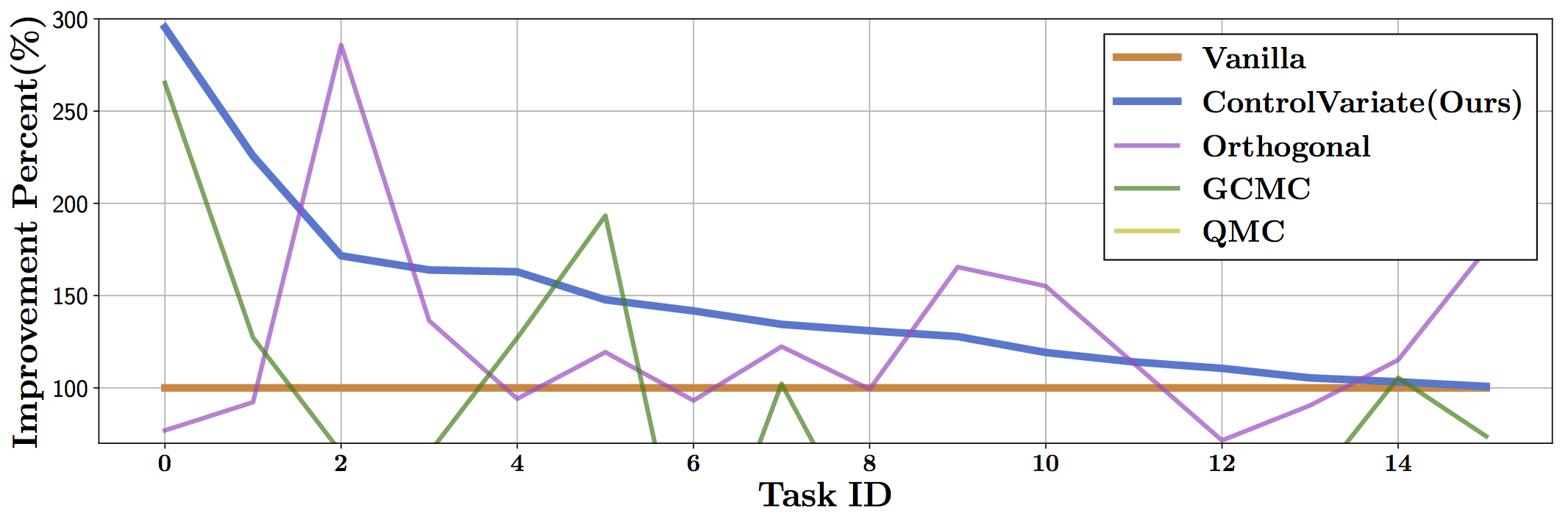}
\caption{Percentile improvement over 16 benchmark tasks. The metric is calculated for each task as $(r_{\text{cv}}- r_{\text{random}}) / (r_{\text{es}} - r_{\text{random}})$, where $r_{\text{cv}}, r_{\text{es}}, r_{\text{random}}$ are the final rewards for the \gls{RL} tasks obtained with our control variate, vanilla \gls{ES}, and random policy methods respectively. We see that our proposal consistently improves over vanilla \gls{ES} for all tasks, and over all compared variance reduction methods for 10 out of 16 tasks. \Cref{sec:exp} provides additional details.}
\label{fig:illustration}
\end{figure*}

\Cref{fig:illustration} summarizes the performance of our proposal over 16 \gls{RL} benchmark tasks. Our method consistently improves over vanilla \gls{ES} baselines and other state-of-the-art, general purpose \gls{MC} variance reduction methods. Moreover, we provide theoretical insight into why our algorithm achieves more variance reduction than orthogonal \gls{ES} \citep{choromanski2018structured} when the policy itself is highly stochastic. (\Cref{sec:new_method} presents a detailed analysis.)

\textbf{Related Work.} Control variates are commonly used to reduce the variance of \gls{MC}-based estimators \citep{ross6277simulation}. In black-box variational inference algorithms, carefully designed control variates can reduce the variance of \gls{MC} gradient updates, leading to faster convergence \citep{paisley2012variational,ranganath2014black}. In \gls{RL}, \gls{PG} methods apply state-dependent baseline functions as control variates \citep{mnih2016,schulman2015,schulman2017}. While action-dependent control variates \citep{gu2016q,liu2017action,grathwohl2017backpropagation,wu2018variance} have been proposed to achieve further variance reduction, \citet{tucker2018mirage} recently showed that the reported performance gains may be due to subtle implementation details, rather than better baseline functions.

To leverage the \gls{MDP} structure in developing a \gls{RL}-specific control variate for \gls{ES}, we derive a gradient estimator for the \gls{ES} objective based on reparameterization \citep{kingma2013} and the \gls{PG} estimator. The control variate is constructed as a difference between two alternative gradient estimators. Our approach is related to a control variate techniques developed for modeling discrete latent variables in variational inference \citep{tucker2017rebar}. The idea is to relax the discrete model into a differentiable one and construct the control variate as the difference between the score function gradient estimator and reparameterized gradient estimator of the relaxed model. We expand on this connection in \Cref{sub:REBAR}.

\section{Policy Optimization in Reinforcement Learning}
\label{sec:rl}

Sequential decision making problems are often formulated as a \gls{MDP}s. Consider an episode indexed by time. At any given time $t\geq0$, an agent is in a state $s_t \in \mathcal{S}$. The agent then takes an action $a_t \in \mathcal{A}$, receives an instant reward $r_t = r(s_t,a_t) \in \mathbb{R}$, and transitions to the next state $s_{t+1} \sim p(\cdot \mid s_t, a_t)$, where $p$ is a distribution determining transitional probabilities. Define the policy  $\pi:\mathcal{S}\mapsto \mathcal{P}(\mathcal{A})$ as a conditional distribution over actions $\mathcal{A}$ given a state $s \in \mathcal{S}$. \Gls{RL} seeks to maximize the expected cumulative rewards over a given time horizon $T$,
\begin{align}
    J^\gamma(\pi)
    &=
    \mathbb{E}_\pi
    \left[
    \sum_{t=0}^{T-1} r_t \gamma^t
    \right],
    \label{eq:rlobj}
\end{align}
where $\gamma \in (0,1]$ is a discount factor and the expectation is with respect to randomized environment and outputs of the policy $\pi$.

Ideally, we would like to work with an infinite horizon and no discount factor. In practice, horizon $T$ is bounded by sample collection \citep{brockman2016} while directly optimizing the undiscounted objective $J^1(\pi)$ admits unusably high variance gradient estimators \citep{schulman2015}. As a result, modern \gls{RL} algorithms tackle the problem through a discount factor $\gamma < 1$, which reduces the variance of the gradient estimators but introduces bias \citep{schulman2015,schulman2017,mnih2016}. At evaluation time, the algorithms are evaluated with finite horizons $T<\infty$ and undiscounted returns $\gamma=1$ \citep{schulman2015,schulman2017,mnih2016}. We follow this setup here as well.

% This can be attributed to several practical issues: (1) though the original MDP can have an infinite horizon, it is only possible to collect finite time steps per episode \citep{brockman2016}; (2) directly optimizing the undiscounted objective $J^1(\pi)$ leads to very high variance of the policy gradient estimators (\textcolor{red}{details below}). Computing gradient estimators through a discounted objective trades bias with variance.

%\sacomment{is $\theta \in \mathbb{R}^\text{some dimension}$? please specify below}

Consider parameterizing the policy as $\pi_\theta$ where $\theta \in \mathbb{R}^d$. The goal is to optimize \Cref{eq:rlobj} with respect to policy parameters. A natural approach is to use exact gradient methods. Regrettably, this objective function does not admit an analytic gradient. Thus, we turn to stochastic gradient techniques \citep{robbins1951stochastic} and seek to construct approximations to the true gradient $g_\theta^\gamma = \nabla_\theta J^\gamma(\pi_\theta)$.

% We begin by introducing methods that approximate the gradient $g_{\theta}$ of \Cref{eq:rlobj}, where a vector $\theta$ encodes the policy $\pi$. ES/PG methods construct gradient approximations $\hat{g}_\theta$ and apply standard gradient ascent updates to conduct optimization (e.g: $\theta \leftarrow \theta +\alpha \hat{g}_\theta$ for some learning rate $\alpha > 0$).

%In this paper, we tackle the finite horizon problem

%Let $\pi^\ast$ denote the optimal policy $\pi^\ast = \arg\max_\pi J^\gamma(\pi)$.

%To tractably search for a sub-optimal policy in the entire policy space, consider parameterizing a policy $\pi_\theta$ with parameter $\theta \in \Theta$. When the function class $\{\pi_\theta,\theta\in \Theta\}$ is sufficiently large, we might expect to find a parameterized policy close to optimal $\pi_{\theta^\ast} \approx \pi^\ast$.

\subsection{Evolution Strategies for Policy Optimization}

\glsreset{ES}
\Gls{ES} \citep{salimans2017evolution} take a black-box optimization approach to maximizing \Cref{eq:rlobj}. To do so, the first step is to ensure that the objective function is differentiable with respect to the policy parameters. To this end, \gls{ES} begin by convolving the original objective $J^\gamma(\pi_\theta)$ with a multivariate isotropic Gaussian distribution of mean $\theta$ and variance $\sigma^2$:
\begin{align}
    F^{\sigma,\gamma}(\theta)
    &=
    \mathbb{E}_{\theta^\prime \sim \mathcal{N}(\theta,\sigma^2 \mathbb{I})}
    \left[
    J^\gamma(\pi_{\theta^\prime})
    \right].
    \label{eq:esobj}
\end{align}

% We refer to Evolution Strategies (ES) as a class of recently revived blackbox optimization algorithms \citep{salimans2017evolution}, where underlying RL task of finding optimal policy is cast as blackbox optimization problem with vectorized policy $\theta$ being an input and an unbiased noisy estimate of $J^\gamma(\pi_\theta)$ (obtained by executing the policy in a given environment) as an output.
% Consider convolving the original objective $J^\gamma(\pi_\theta)$ with an multivariate isotropic Gaussian distribution with mean $\theta$ and  diagnonal covariance matrix $\sigma^2 \mathbb{I}$:
% \begin{align}
%     F^{\sigma,\gamma}(\theta) = \mathbb{E}_{\theta^\prime \sim \mathcal{N}(\theta,\sigma^2 \mathbb{I})} [ J^\gamma(\pi_{\theta^\prime})].
%     \label{eq:esobj}
% \end{align}

\gls{ES} maximize this smooth objective as a proxy to maximizing the  original objective $J^\gamma(\pi_\theta)$. The convolved objective $F$ enjoys the advantage of being differentiable with respect to the policy. In the limit $\sigma \rightarrow 0$, an optimal point of $F^{\sigma,\gamma}(\theta)$ is also optimal with respect to $J^\gamma(\pi_\theta)$. The next step is to derive a gradient of \Cref{eq:esobj}. Consider the score function gradient,
\begin{align}
   \nabla_\theta F^{\sigma,\gamma}(\theta)
   &=
   \mathbb{E}_{\theta^\prime \sim \mathcal{N}(\theta,\sigma^2 \mathbb{I})}
   \left[
   J^\gamma(\pi_{\theta^\prime}) \frac{\theta^\prime-\theta}{\sigma^2}
   \right].
\label{eq:esgradtrue}
\end{align}
This gradient can be computed by sampling $\theta_i^\prime \sim \mathcal{N}(\theta,\sigma^2\mathbb{I})$ and computing unbiased estimates of each $J^{\gamma}(\pi_{\theta_i^\prime})$ using a single roll-out trajectory of $\pi_{\theta_i^\prime}$ in the environment. The resulting score function gradient estimator has the following form:
\begin{align}
    \hat{g}^{\textsc{es},\gamma}_\theta = \frac{1}{N} \sum_{i=1}^N J^\gamma(\pi_{\theta_i^\prime}) \frac{\theta_i^\prime - \theta}{\sigma^2}.
    \label{eq:esgrad}
\end{align}
This gradient estimator is biased with respect to the original objective. However, in practice this bias does not hinder optimization; on the contrary, the smoothed objective landscape is often more amenable to gradient-based optimization \citep{leordeanu2008smoothing}. We also make clear that though the \gls{ES} gradient is defined for any $\gamma \in (0,1]$, in practice parameters are updated with the gradient of the undiscounted objective $\hat{g}_\theta^{\text{ES},1}$ \citep{salimans2017evolution,mania2018simple,choromanski2018structured}.

\subsection{Policy Gradient Methods for Policy Optimization}
\glsreset{PG}
\Gls{PG} \citep{sutton1999} methods take a different approach. Instead of deriving the gradient through a parameter level perturbation as in \gls{ES}, the core idea of \gls{PG} is to leverage the randomness in the policy itself. Using a standard procedure from stochastic computational graphs \citep{schulman2015gradient}, we compute the gradient of \Cref{eq:rlobj} as follows
\begin{align}
    \nabla_\theta J^\gamma(\pi_\theta)
    &=
    \mathbb{E}_{\pi_\theta}
    \left[
    \sum_{t=0}^{T-1}
    \left(
    \sum_{t^\prime = t}^{T-1} r_{t^\prime}
    \right)
    \gamma^t \nabla_\theta \log \pi_\theta(a_t|s_t)
    \right].
    \label{eq:pg}
\end{align}
Unbiased estimators $\hat{g}^{\textsc{pg},\gamma}_\theta$ of this gradient can be computed using sampling as above for the \gls{ES} method. In practice, the sample estimate of \Cref{eq:pg} often has large variance which destabilizes the updates. To alleviate this issue, one convenient choice is to set $\gamma < 1$ so that the long term effects of actions are weighted down exponentially. This reduces the variance of the estimator, but introduces bias with respect to the original undiscounted objective $J^1(\pi_\theta)$.

\Gls{ES} and \gls{PG} are two alternative methods for deriving gradient estimators with respect to the policy parameters. On an intuitive level, these two methods complement each other for variance reduction: \gls{PG} leverages the \gls{MDP} structure and achieves lower variance when the policy is stochastic; \gls{ES} derives the gradient by injecting noise directly into the parameter space and is characterized by lower variance when the policy itself is near-deterministic. Our goal in this paper is to develop a single estimator that benefits from both approaches. We formalize this intuition in the next section.

%\textcolor{red}{alp: we need a transition: Summarize the benefits of \gls{PG} and \gls{ES} and why we are motivated to combine them. I will edit this transition once you write it.}

\section{Variance Reduction via Structured Control Variates}
\label{sec:new_method}
We seek a control variate for the \gls{ES} gradient estimator in \Cref{eq:esgrad}. Recall that this gradient is with respect to a smoothed objective: $F^{\sigma,\gamma}(\theta)$ with $\gamma \in (0,1]$.

\subsection{Reparameterized Gradients of the Smoothed Objective}
The \gls{ES} gradient estimator in \Cref{eq:esgrad} leverages the derivative of the logarithm. We can also apply the reparameterization technique \citep{kingma2013} to the distribution $\theta^\prime \sim \mathcal{N}(\theta,\sigma^2\mathbb{I})$ to obtain:
\begin{align}
    \nabla_\theta F^{\sigma,\gamma}(\theta) &= \nabla_\theta \mathbb{E}_{\theta^\prime \sim \mathcal{N}(\theta,\sigma^2\mathbb{I})} [J^\gamma(\pi_{\theta^\prime})] \nonumber \\
    &= \nabla_\theta \mathbb{E}_{\epsilon \sim \mathcal{N}(0,\mathbb{I})} [J^\gamma(\pi_{\theta +\epsilon \cdot \sigma})] \nonumber \\
    &=  \mathbb{E}_{\epsilon \sim \mathcal{N}(0,\mathbb{I})} [\nabla_{\theta + \epsilon \cdot \sigma}J^\gamma(\pi_{\theta +\epsilon \cdot \sigma})], \label{eq:reparameterizedes}
\end{align}
where $\nabla_{\theta +\epsilon\cdot\sigma} J^\gamma(\pi_{\theta+\epsilon\cdot\sigma})$ can be computed by \gls{PG} estimators for the discounted objective (\ref{eq:pg}). To estimate (\ref{eq:reparameterizedes}), we sample $\epsilon_i \sim \mathcal{N}(0,\sigma^2\mathbb{I})$ and construct perturbed policies $\theta_i^\prime = \theta + \epsilon_i \cdot\sigma$. Then an unbiased estimate $\hat{g}^{\textsc{pg},\gamma}_{\theta + \epsilon_i \cdot\sigma}$ of the policy gradient $\nabla_{\theta + \epsilon_i \cdot\sigma} J^\gamma (\pi_{\theta + \epsilon_i \cdot\sigma}) $ can be computed from a single rollout trajectory using $\pi_{\theta + \epsilon_i\cdot\sigma}$. Finally the reparameterized gradient is computed by averaging:
\begin{align}
    \hat{g}^{\textsc{re},\gamma}_\theta = \frac{1}{N} \sum_{i=1}^N  \hat{g}^{\textsc{pg},\gamma}_{\theta + \epsilon_i \cdot\sigma}.
    \label{eq:reesgrad}
\end{align}

\subsection{Evolution Strategies with Structured Control Variates}
For the discounted objective $F^{\sigma,\gamma}(\theta)$ we have two alternative gradient estimators. One is constructed using the score function gradient estimator (see: \Cref{eq:esgrad}). The other uses the re-parameterization techqniue along with policy gradient estimators (see: \Cref{eq:reesgrad}). Combining these two estimators with the vanilla \gls{ES} gradient for the undiscounted objective $\hat{g}^{\textsc{es},1}_\theta$, we get:
\begin{align}
    \hat{g}^{\textsc{cv}}_\theta = \hat{g}^{\textsc{es},1}_\theta + \eta \odot (\hat{g}^{\textsc{es},\gamma}_\theta - \hat{g}^{\textsc{re},\gamma}_\theta),
    \label{eq:escv}
\end{align}
where $\eta$ is a vector of same dimension of $\theta$ and $\odot$ denotes an element-wise product. This scaling parameter $\eta$ controls the relative importance of the two terms in (\ref{eq:escv}). As discussed below, we can adapt the discount factor $\gamma$ and the scaling parameter $\eta$ to minimize the variance over time.

 \paragraph{Discount factor $\gamma$.}
  As in \citet{choromanski2018structured}, for a vector $g\in \mathbb{R}^n$, we define its variance as the sum of its component variances $\mathbb{V}[g] \coloneqq \sum_{i=1}^n \mathbb{V}[g_i]$. We then adapt the discount factor $\gamma \leftarrow \gamma - \alpha_\gamma \nabla_\gamma \mathbb{V}[\hat{g}_\theta^{\textsc{cv}}]$ for some learning rate $\alpha_\gamma > 0$. Since $\mathbb{E}[\hat{g}_\theta^{\textsc{cv}}]$ does not depend on $\gamma$, we have equivalently $\nabla_\gamma \mathbb{V}[\hat{g}_\theta^{\textsc{cv}}] = \nabla_\gamma \mathbb{E}[(\hat{g}_\theta^{\textsc{cv}})^2]$. The gradient $\nabla_\gamma \mathbb{E}[(\hat{g}_\theta^{\textsc{cv}})^2]$ can be itself estimated using backpropagation on mini-batches but this tends to be unstable because each term in (\ref{eq:pg}) involves $\gamma^t$. Alternatively, we build a more robust estimator of $\nabla_{\gamma} \mathbb{E}[(\hat{g}_\theta^{\textsc{cv}})^2]$ using \gls{ES}: in particular, sample $\epsilon_i \sim \mathcal{N}(0,1)$ and let $v_i$ be the evaluation of $\mathbb{E}[(\hat{g}_\theta^{\textsc{cv}})^2]$ under $\gamma +\sigma_\gamma \epsilon_i$ for some $\sigma_\gamma > 0$. The gradient estimator for $\gamma$ is $\hat{g}_\gamma = \frac{1}{N}\sum_{i=1}^N v_i \frac{\epsilon_i}{\sigma_\gamma}$.

  Though the full estimator (\ref{eq:escv}) is defined for all discount factors $\gamma \in (0,1]$, in general we find it better to set $\gamma < 1$ to stablize the \gls{PG} components of the control variate.

 \paragraph{Coefficient $\eta$.} Since $\eta$ is a vector with the same dimensionality as $\theta$, we can update each component of $\eta$ to reduce the variance of each component of $\hat{g}_{\theta}^{\textsc{cv}}$. Begin by computing, $\nabla_\eta \mathbb{V}[\hat{g}_{\theta}^{\textsc{cv}}]$ as follows:
 \begin{align}
     \nabla_\eta \mathbb{V}
     \bigg[
     \hat{g}_{\theta}^{\textsc{cv}}
     \bigg]
     &=
     2 \eta \odot \mathbb{E}
     \bigg[
     \big(
     \hat{g}^{\textsc{es},\gamma}_\theta - \hat{g}^{\textsc{re},\gamma}_\theta
     \big)^2
     \bigg]
     + \nonumber \\
     & 2 \mathbb{E}
     \bigg[
     \big(
     \hat{g}^{\textsc{es},\gamma}_\theta - \hat{g}^{\textsc{re},\gamma}_\theta
     \big)
     \odot \hat{g}_{\theta}^{\textsc{es},1}
     \bigg].
     \label{eq:etagrad}
 \end{align}
Then, estimate this gradient using \gls{MC} sampling. Finally, adapt $\eta$ by running online gradient descent: $\eta \leftarrow \eta - \alpha_\eta \nabla_\eta \mathbb{V}[\hat{g}_\theta^{\textsc{cv}}]$ with some $\alpha_\eta > 0$.

\paragraph{Practical considerations.} Certain practical techniques can be applied to stabilize the \gls{ES} optimization procedure. For example, \citet{salimans2017evolution} apply a centered rank transformation to the estimated returns $J^\gamma(\pi_{\theta_i^\prime})$ to compute the estimator of the gradient in \Cref{eq:esgrad}. This transformation is compatible with our proposal. The construction becomes $\hat{g}_\theta^{\textsc{cv}} = \hat{g}_\theta  +\eta (\hat{g}^{\textsc{es},\gamma}_\theta - \hat{g}^{\textsc{re},\gamma}_\theta)$ where $\hat{g}_\theta$ can be computed through the rank transformation.

\paragraph{Stochastic policies.} While many prior works \citep{salimans2017evolution,mania2018simple,choromanski2018structured,rowland2018geometrically} focus on deterministic policies for continuous action spaces, our method targets stochastic policies, as required by the \gls{PG} computation. Estimating \gls{PG} for a deterministic policy requires training critic functions and is in general biased \citep{silver2014}. We leave the investigation of deterministic policies to future work.

%\footnote{We note that this is not a big restriction, since stochastic policy can be applied to both continuous and discrete action domain.}

\subsection{Relationship to REBAR}
\label{sub:REBAR}

REBAR \citep{tucker2017rebar} considers variance reduction of gradient estimators for probabilistic models with discrete latent variables. The discrete latent variable model has a \emph{relaxed model} version, where the discrete sampling procedure is replaced by a differentiable function with reparameterized noise. This relaxed model usually has a temperature parameter $\tau$ such that when $\tau \rightarrow 0$ the relaxed model converges to the original discrete model. To optimize the discrete model, the baseline approach is to use score function gradient estimator, which is unbiased but has high variance. Alternatively,  one could use the reparameterized gradient through the relaxed model, which has lower variance, but the gradient is biased for finite $\tau > 0$ \citep{jang2016categorical,maddison2016concrete}. The bias and variance of the gradient through the relaxed model is controlled by $\tau$.  REBAR proposes to use the difference between the score function gradient and reparameterized gradient of the relaxed model as a control variate. Their difference has expectation zero and should be highly correlated with the reinforced gradient of the original discrete model, leading to potentially large variance reduction.

A similar connection can be found in \gls{ES} for \gls{RL} context. We can interpret the non-discounted objective, namely $F^{\sigma,\gamma}(\pi_\theta)$ with $\gamma =1$, as the original model which gradient we seek to estimate. When $\gamma < 1$, we have the relaxed model which gradient becomes biased but  has lower variance (with respect to the non-discounted objective). Similar to REBAR, our proposal is to construct the score function gradient (\gls{ES} estimator) $\hat{g}_\theta^{\textsc{es},\gamma}$ and reparameterized gradient (\gls{PG} estimator) $\hat{g}_\theta^{\textsc{re},\gamma}$ of the general discounted objective $\gamma < 1$ (relaxed model), such that their difference $\hat{g}_\theta^{\textsc{es},\gamma}-\hat{g}_\theta^{\textsc{re},\gamma}$ serves as a control variate. The variance reduction from REBAR applies here, $\hat{g}_\theta^{\textsc{es},\gamma}-\hat{g}_\theta^{\textsc{re},\gamma}$  should be highly correlated with $\hat{g}_\theta^{\textsc{es},1} $, which leads to effective variance reduction.

\subsection{How much variance reduction is possible?}

How does the variance reduction provided by control variate compare to that of general purpose methods, such as orthogonalization \citep{choromanski2018structured}? In this section, we build on a simple example to illustrate the different variance reduction properties of these approaches. Recall that we define the variance of a vector $g\in \mathbb{R}^d$ as the sum of the variance of its components $\mathbb{V}[g] = \sum_{i=1}^d \mathbb{V}[g_i]$ following notation from prior literature \citep{choromanski2018structured,rowland2018geometrically}.

Consider a one-step \gls{MDP} problem where the agent takes only one action $a$ and receives a reward $r(a) = \alpha^T a$ for some $\alpha \in \mathbb{R}^d$. We choose the reward function to be linear in $a$, as a local approximation to a potentially nonlinear reward function landscape. Let the policy be a Gaussian with mean $\mu$ and diagonal covariance matrix $\Sigma = \sigma_2^2\mathbb{I}$ with fixed $\sigma_2$. Here the policy parameter contains only the mean $\theta = \mu$. The \gls{RL} objective is $J(\pi_\mu) = \mathbb{E}_{a\sim \mathcal{N}(\mu,\Sigma)}[r(a)]$. To compute the gradient, \gls{ES} smoothes the objective with the Gaussin $\mathcal{N}(\mu,\sigma_1^2\mathbb{I})$ for a fixed $\sigma_1$. While vanilla \gls{ES} generates i.i.d.~perturbations $\epsilon_i$, orthogonal \gls{ES} couples the perturbations such that ${\epsilon_i^\prime}^T \epsilon_j^\prime = 0,i\neq j$.

Denote $d$ as the dimensionality of the parameter space $\mu \in \mathbb{R}^d$ and let $\rho = \sfrac{\sigma_2}{\sigma_1}$. Let $\hat{J}(\pi_{\mu1})$ be a one-sample noisy estimate of $J(\pi_\mu)$. Recall that \gls{ES} gradient takes the form $\hat{g}_\mu^{\textsc{es}} = \sfrac{1}{N}\sum_{i=1}^N \hat{J}(\pi_{\mu+\sigma_1\epsilon_i}) \: \sfrac{\epsilon_i}{\sigma_1}$ (\cref{eq:esgrad}). The orthogonal \gls{ES} gradient takes the same form, but with orthogonal perturbations
$\hat{g}_\mu^{\text{ort}} = \sfrac{1}{N}\sum_{i=1}^N \hat{J}(\pi_{\mu+\sigma_1\epsilon_i^\prime}) \: \sfrac{\epsilon_i^\prime}{\sigma_1}$. Finally, the \gls{ES} with control variate produces an estimator of the form (\ref{eq:escv}). We are now ready to provide the following theoretical result.

\begin{theorem}
\label{thm:varianceratio} In the one-step \gls{MDP} described above, the
ratio of the variance of the orthogonal \gls{ES} to the 
variance of the vanilla \gls{ES}, and the corresponding ratio for the control variate \gls{ES} satisfy respectively:
\begin{align}
    \ \ \ \ \ \ \ \ \ \frac{\mathbb{V}[\hat{g}_{\mu}^{\text{ort}}]}{\mathbb{V}[\hat{g}_{\mu}^{\textsc{es}}]} &= 1 - \frac{N-1}{(1+\rho^2)d + 1}, \nonumber \\  \ \  \frac{\mathbb{V}[\hat{g}_{\mu}^{\textsc{cv}}]}{\mathbb{V}[\hat{g}_{\mu}^{\textsc{es}}]} &\leq 1 - \frac{\rho^2[d((1+\rho^2) - 4]}{[(1+\rho^2)d + 1](1+\rho^2)}. \nonumber
\end{align}
As a result, there exists a threshold $\rho_0$ such that when $\rho \geq \rho_0$, we always have $\mathbb{V}[\hat{g}_{\mu}^{\textsc{cv}}] \leq \mathbb{V}[\hat{g}_{\mu}^{\text{ort}}]$. (See: Appendix for details). Importantly, when $d$ is large enough, we have $\rho_0 \rightarrow \sqrt{\sfrac{N}{d}}$.
\end{theorem}

Some implications of the above theorem: \textbf{(1)} For orthogonal \gls{ES}, the variance reduction depends explicitly on the sample size $N$. In cases where $N$ is small, the variance gain over vanilla \gls{ES} is not significant. On the other hand, $\mathbb{V}[\hat{g}_\mu^{\textsc{cv}}]$ depends implicitly on $N$ because in practice $\eta^\ast$ is approximated via gradient decent and large sample size $N$ leads to more stable updates; \textbf{(2)} The threshold $\rho_0$ is useful in practice. In high-dimensional applications where sample efficiency is important, we have large $d$ and small $N$. This implies that for a large range of the ratio $\rho = \frac{\sigma_2}{\sigma_1} \geq \rho_0$, we could expect to achieve more variance reduction than orthogonal \gls{ES}. \textbf{(3)} The above derivation is based on the simplification of the general multi-step \gls{MDP} problem. The practical performance of the control variate can also be influenced by how well $\eta^\ast$ is estimated. Nevertheless, we expect this example to provide some guideline as to how the variance reduction property of \gls{ES} with control variate depends on $\rho$ and $N$, in contrast to orthogonal \gls{ES}; \textbf{(4)} The theoretical guarantee for variance reduction of orthogonal \gls{ES}  \citep{choromanski2018structured} relies on the assumption that $\hat{J}(\pi_{\theta^\prime})$ can be simulated without noise \footnote{The variance reduction proof can be extended to cases where $\hat{J}(\pi_{\theta^\prime})$ has the same level of independent noise for all $\theta^\prime$.}, which does not hold in practice. In fact, in \gls{RL} the noise of the reward estimate heavily depends on the policy $\pi_{\theta^\prime}$ (intuitively the more random the policy is, the more noise there is in the estimate). On the other hand, \gls{ES} with control variate depends less on such assumptions but rather relies on finding the proper scalar $\eta$ using gradient descent. We will see in the experiments that this latter approach reliably improves upon the \gls{ES} baseline.

\section{Experiments}
\label{sec:exp}

In the experiments, we aim to address the following questions: \textbf{(1)}: Does the control variate improve downstream training through variance reduction? \textbf{(2)}: How does it compare with other recent variance reduction techniques for \gls{ES}?

To address these questions, we compare our control variate to general purpose variance reduction techniques on a wide range of high-dimensional \gls{RL} tasks with continuous action space.
\begin{description}
    \item[Antithetic sampling:] See \citet{salimans2017evolution}. Let $\epsilon_i,1\leq i\leq N$ be the set of perturbation directions. Antithetic sampling perturbs the policy parameter with a set of antithetic pairs $(\epsilon_i,\epsilon_i^\prime),1\leq i\leq N$ where $\epsilon_i^\prime = -\epsilon_i$.
    \item[Orthogonal directions (ORTHO):] See \citet{choromanski2018structured}. The set of perturbations $\epsilon_i,1\leq i\leq N$ applied to the policy parameter are generated such that they have the same marginal Gaussian distributions but are orthogonal to each other $\epsilon_i^T \epsilon_j = 0, i\neq j$.
    \item[Geometrically coupled MC sampling (GCMC):] See \citet{rowland2018geometrically}. For each antithetic pair $(\epsilon_i,\epsilon_i^\prime)$,  GCMC couples their length such that $F_R(\Vert\epsilon_i\Vert)+F_R(\Vert\epsilon_i^\prime\Vert) = 1$ where $F_R$ is the CDF of the norm of a standard Gaussian with the same dimension as $\epsilon_i$.
    \item[Quasi Monte-Carlo (QMC):] See \citet{choromanski2018structured,rowland2018geometrically}. QMC first generates a low-discrepancy Halton sequence $\{r_i\}_{i=1}^N$ in $[0,1]^d$ with $N$ elements where $d$ is the dimension of parameter $\theta$. Then apply the inverse CDF $F_g^{-1}$ of a standard univariate Gaussian elementwise to the sequence $\epsilon_i = F_g^{-1}(r_i)$ to generate perturbation vectors.
\end{description}
We find that our \gls{RL}-specific control variate achieves outperforms these general purpose variance reduction techniques. In addition to the above baseline comparison, we also compare CV with more advanced policy gradient algorithms such as TRPO / PPO \citep{schulman2015,schulman2017}, as well as deterministic policy $+$ \gls{ES} baseline \citep{mania2018simple}.

\textbf{Implementation details.} Since antithetic sampling is the most commonly applied variance reduction method, we combine it with the control variate, ORTHO, GCMC and QMC. The policy $\pi_\theta$ is parameterized as a neural network with 2 hidden layers each with 32 units and $\text{relu}$ activation function. The output is a vector $\mu_\theta \in \mathbb{R}^K$ used as a Gaussian mean, with a separately parameterized diagonal vector $\sigma_0 \in \mathbb{R}^K$ independent of state. The action is sampled the Gaussian  $a\sim \mathcal{N}(\mu_\theta(s),\text{diag}(\sigma^2))$. The backpropagation pipeline is implemented with Chainer \citep{tokui2015chainer}. The learning rate is $\alpha = 0.01$ with Adam optimizer \citep{kingma2014adam}, the perturbation standard deviation $\sigma = 0.02$. At each iteration we have $N=5$ distinct perturbations $\epsilon_i$ ($2N$ samples in total due to antithetc sampling). For the control variate (\ref{eq:escv}), the discount factor is initialized to be $\gamma = 0.99$ and updated with \gls{ES}, we introduce the details in the Appendix. The control variate scaling factor $\eta$ is updated with learning rate selected from $\alpha_\eta \in \{10^{-3},10^{-4},10^{-5}\}$. As commonly practiced in prior works \citep{salimans2017evolution,mania2018simple,choromanski2018structured}, in order to make the gradient updates less sensitive to the reward scale, returns are normalized before used for computing the gradients. We adopt this technique and discuss the details in the Appendix. Importantly, we do not normalize the observations (as explained in \citep{mania2018simple}) to avoid additional biasing of the gradients.

\textbf{Benchmark tasks and baselines.} To evaluate how variance reduction impacts downstream policy optimization, we train  neural network policies over a wide range of high-dimensional continuous control tasks, taken from OpenAI gym \citep{brockman2016}, Roboschool \citep{schulman2017} and DeepMind Control Suites \citep{tassa2018deepmind}. We introduce their details below. We also include a LQR task suggested in \citep{mania2018simple} to test the stability of the gradient update for long horizons ($T=2000$). Details of the tasks are in the Appendix. The policies are trained with five variance reduction settings: Vanilla \gls{ES} baseline; \gls{ES} with orthogonalization (ORTHO); \gls{ES} with GCMC (GCMC); \gls{ES} with Quasi-\gls{MC} (QMC); and finally our proposed \gls{ES} with control variate (CV).

\begin{figure*}[h]
\centering
\subfigure[\textbf{LQR}]{\includegraphics[width=.23\linewidth]{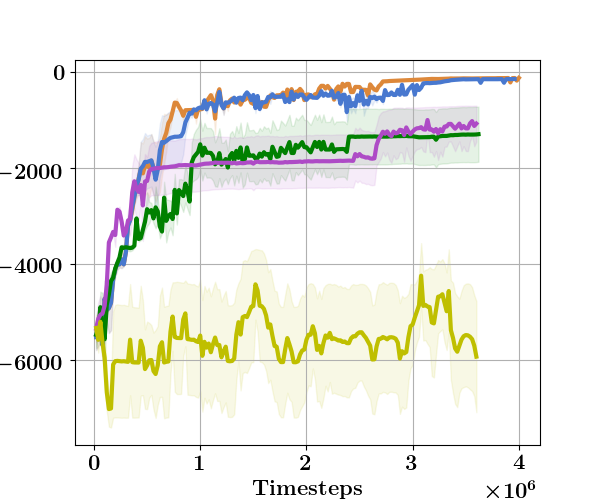}}
\subfigure[\textbf{Swimmer}]{\includegraphics[width=.23\linewidth]{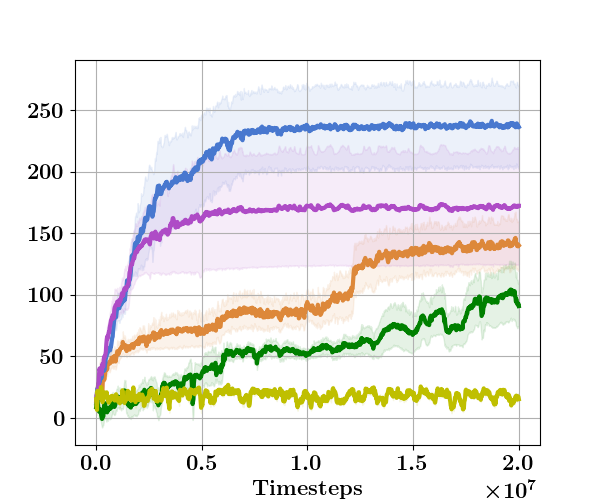}}
\subfigure[\textbf{HalfCheetah}]{\includegraphics[width=.23\linewidth]{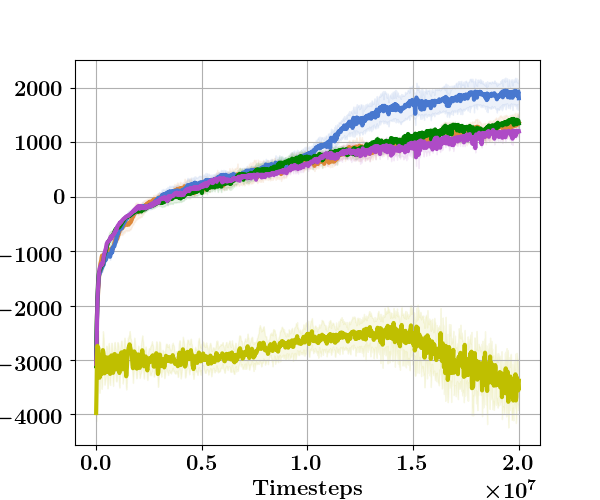}}
\subfigure[\textbf{Walker}]{\includegraphics[width=.23\linewidth]{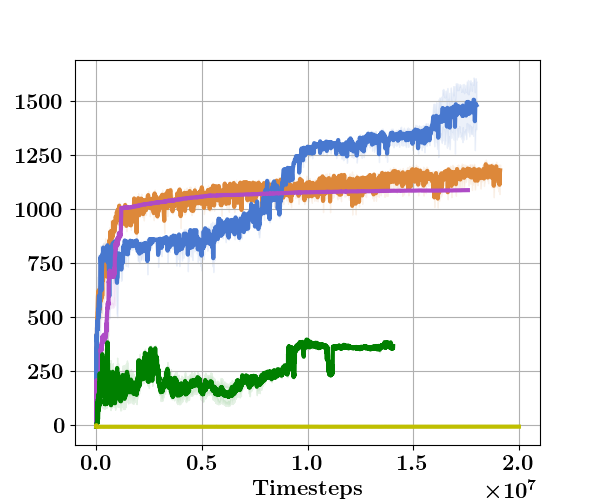}}
\subfigure[\textbf{RoboschoolPong}]{\includegraphics[width=.23\linewidth]{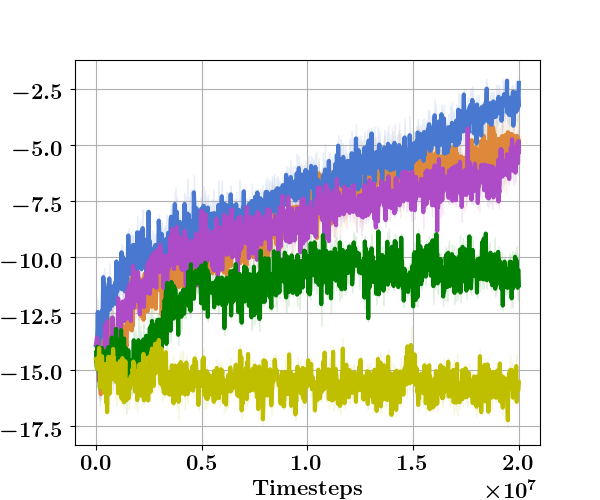}}
\subfigure[\textbf{RoboschoolCheetah}]{\includegraphics[width=.23\linewidth]{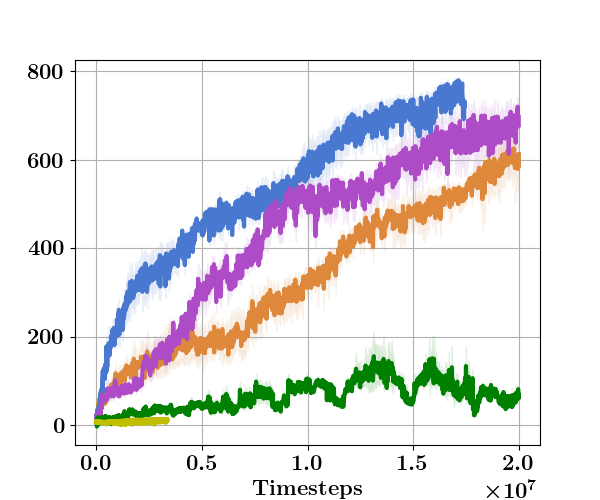}}
\subfigure[\textbf{BipedalWalker}]{\includegraphics[width=.23\linewidth]{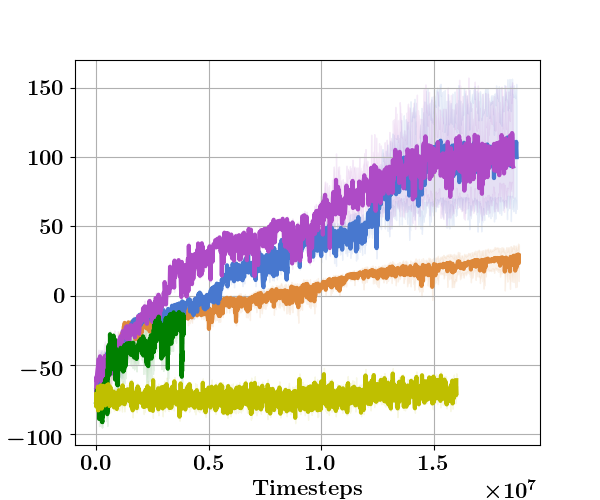}}
\subfigure[\textbf{Swingup (DM)}]{\includegraphics[width=.23\linewidth]{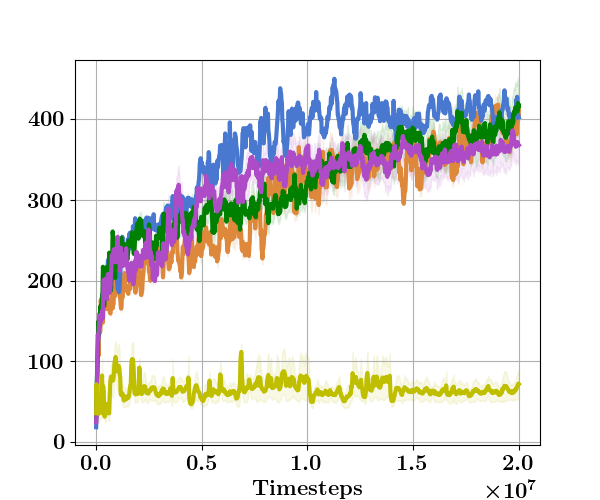}}
\subfigure[\textbf{TwoPoles (DM)}]{\includegraphics[width=.23\linewidth]{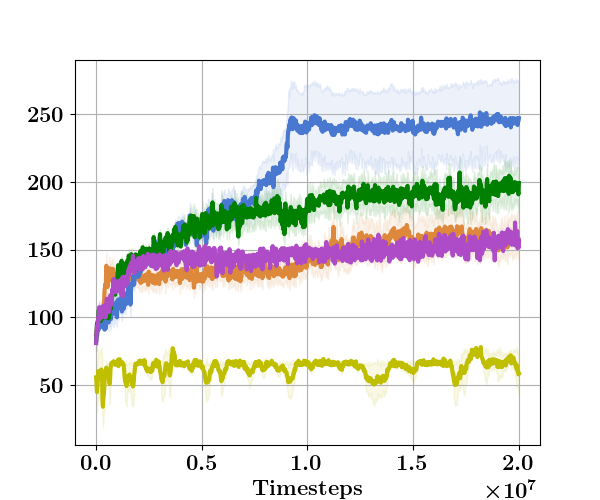}}
%\subfigure[\textbf{Balance (DM)}]{\includegraphics[width=.23\linewidth]{plots/es_long_dmcartpolebalance.png}}
%\subfigure[\textbf{Pendulum (DM)}]{\includegraphics[width=.23\linewidth]{plots/es_long_dmpendulumswingup.png}}
\subfigure[\textbf{CheetahRun (DM)}]{\includegraphics[width=.23\linewidth]{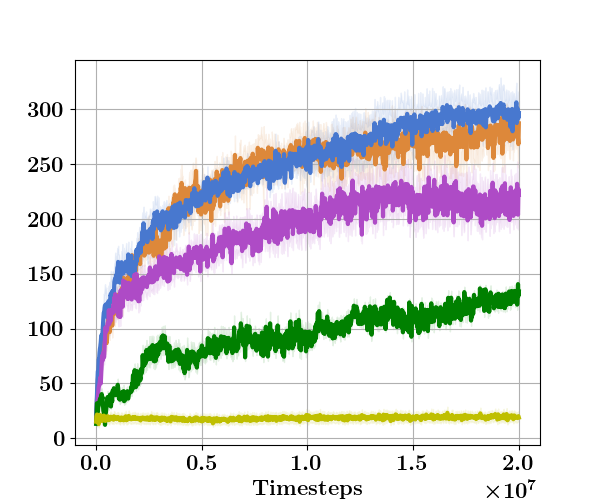}}
\subfigure[\textbf{HopperStand (DM)}]{\includegraphics[width=.23\linewidth]{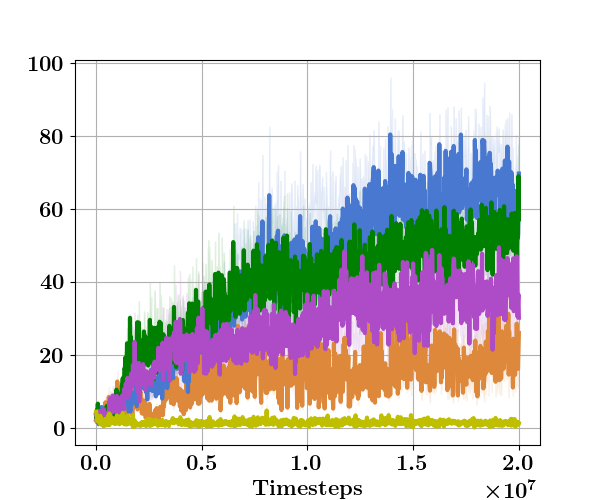}}
%\subfigure[\textbf{HopperHop (DM)}]{\includegraphics[width=.23\linewidth]{plots/esrank_long_dmhopperhop.png}}
%\subfigure[\textbf{AntWalk(DM)}]{\includegraphics[width=.23\linewidth]{plots/esrank_long_dmantwalk.png}}
\subfigure[\textbf{AntEscape(DM)}]{\includegraphics[width=.23\linewidth]{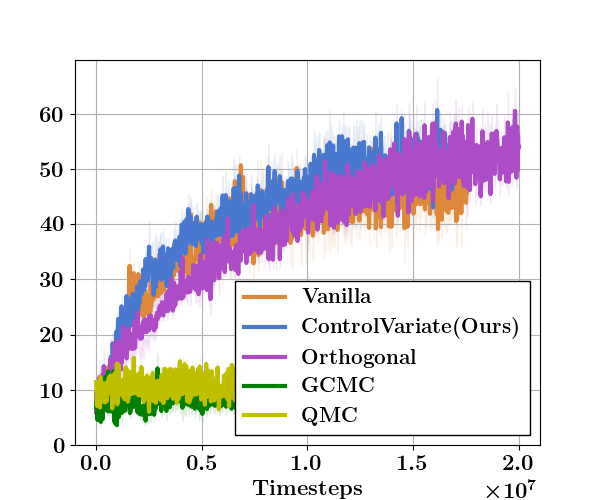}}
\caption{\small{Training performance on Continuous Control Benchmarks: Swimmer, HalfCheetah, CartPole $+$ $\{\text{Swingup},\text{TwoPoles},\text{Balance}\}$, Pendulum Swingup, Cheetah Run and Hopper. Tasks with DM stickers are from the DeepMind Control Suites. We compare five alternatives: baseline ES (orange), CV (blue, ours), ORTHO (marron), GCMC (green) and QMC (yellow). Each task is trained for $2 \cdot 10^7$ time steps (LQR is trained for $4 \cdot 10^6$ steps) and the training curves show the mean $\pm$ std cumulative rewards across 5 random seeds.}}
\label{figure:objective}
\end{figure*}

\paragraph{Results.}
In each subplot of Figure \ref{figure:objective}, we present the learning curves of each variance reduction technique, with average performance over 5 random seeds and the shaded areas indicate standard deviations. 

We make several observations regarding each variance reduction technique: (1) Though ORTHO and GCMC significantly improve the learning progress over the baseline \gls{ES} under certain settings (e.g. ORTHO in Swimmer and GCMC), their improvement is not very consistent. In certain cases, adding such techniques even makes the performance worse than the baseline \gls{ES}. We speculate that this is because the variance reduction conditions required by these methods are not satisfied, e.g. the assumption of noiseless estimate of returns. Overall, ORTHO is more stable than GCMC and QMC; (2) QMC performs poorly on most tasks. We note that similar results have been reported in \citep{rowland2018geometrically} where they train a navigation policy using QMC and show that the agent does not learn at all. We speculate that this is because the variance reduction achieved by QMC is not worth the bias in the \gls{RL} contexts; (3) No variance reduction technique performs uniformly best across all tasks. However, CV performs the most consistently and achieves stable gains over the vanilla \gls{ES} baseline, while other methods can underperform the vanilla \gls{ES} baseline.

To make clear the comparison of final performance, we record the final performance (mean $\pm$ std) of all methods in Table \ref{table:full}. Best results across each task are highlighted in bold font. For a  fair presentation of the results, in cases where multiple methods achieve statistically similar performance, we highlight all such methods. CV consistently achieves the top results across all reported tasks.

\begin{table*}[t]
\caption{Final performance on benchmark tasks. The policy is trained for a fixed number of steps on each task. The result is $\text{mean} \pm \text{std}$ across 5 random seeds. The best results are highlighted in bold font. We highlight multiple methods if their results cannot be separated ($\text{mean} \pm \text{std}$ overlap). CV (ours) achieves consistent gains over the baseline and other variance reduction methods. We also include a \gls{PG} baseline.}
\vskip 0.1in
\begin{center}
\footnotesize
\begin{sc}
\begin{tabular}{ C{0.9in} *6{C{.8in}}}\toprule[1.5pt]
\bf Tasks & \bf Vanilla ES  & \bf Orthogonal & \bf GCMC & \bf QMC & \bf CV (Ours) & \bf PG \\\midrule
LQR      &  $-176 \pm 12$  & $-1337 \pm 573$ & $-1246 \pm 502$ & $-5634 \pm 1059$ &  $\mathbf{-143 \pm 4}$ & $-7243 \pm 275$ \\
Swimmer      &  $141 \pm 20$  & $171 \pm 47$ & $94 \pm 19$ & $16 \pm 2$ &  $\mathbf{237 \pm 33}$ & $-132 \pm 5$\\
HalfCheetah & $1339 \pm 178$ & $1185 \pm 76$ & $1375  \pm 58$ & $-3466 \pm 338$ & $\mathbf{1897 \pm 232}$ & $-180 \pm 4$  \\
Walker & $1155 \pm 34$ & $1087 \pm 1$ & $360 \pm 4$ & $6 \pm 0$ & $\mathbf{1476 \pm 112}$ & $282 \pm 25$ \\
Pong(R) & $-5.0 \pm 0.8$ & $-5.5 \pm 0.3$ & $-10.6 \pm 0.4$ & $-15.6 \pm 0.3$ & $\mathbf{-3.0 \pm 0.3}$ & $-17 \pm 0.2$\\
HalfCheetah(R) & $\mathbf{595 \pm 42}$ & $\mathbf{685 \pm 34}$ & $68 \pm 8$ & $11 \pm 2$ & $\mathbf{709 \pm 16}$ & $12 \pm 0$\\
BipedalWalker & $25 \pm 9$ & $\mathbf{107 \pm 31}$ & $-19 \pm 5$ & $-70 \pm 3$ & $\mathbf{105 \pm 40}$ & $-82 \pm 12$ \\
Cheetah(DM) & $281\pm 15$ & $217 \pm 15$ & $129 \pm 4$ & $18 \pm 5$ & $\mathbf{296 \pm 15}$ & $25 \pm 6$ \\
Pendulum(DM) & $20 \pm 3$ & $\mathbf{54 \pm 17}$ & $25 \pm 8$ & $11 \pm 2$ & $\mathbf{43 \pm 1}$ & $3 \pm 1$ \\
TwoPoles(DM) & $159 \pm 13$ & $158 \pm 2$ & $196 \pm 12$ & $62 \pm 12$ & $\mathbf{245 \pm 29}$ & $14 \pm 1$ \\
Swingup(DM) & $\mathbf{394 \pm 15}$ & $\mathbf{369 \pm 22}$ & $\mathbf{414 \pm 31}$ & $67 \pm 14$ & $\mathbf{406 \pm 26}$ & $55 \pm 10$ \\
Balance(DM) & $692 \pm 57$ & $771 \pm 41$ & $\mathbf{995 \pm 1}$ & $223 \pm 32$ & $\mathbf{847 \pm 71}$ & $401 \pm 12$\\
HopperHop(DM) & $\mathbf{5.7 \pm 2.1}$ & $\mathbf{6.8 \pm 0.7}$ & $0.3 \pm 0.1$ & $0.0 \pm 0.0$ & $\mathbf{6.5 \pm 1.5}$ & $0.1 \pm 0.0$ \\
Stand(DM) & $21 \pm 5$ & $36 \pm 10$ & $\mathbf{54 \pm 4}$ & $1.0 \pm 0.2$ & $\mathbf{60 \pm 11}$ & $0.5 \pm 0.1$\\
AntWalk(DM) & $\mathbf{200 \pm 19}$ & $\mathbf{234 \pm 10}$ & $82 \pm 11$ & $133 \pm 9$ & $\mathbf{239 \pm 10}$ & $100 \pm 11$ \\
AntEscape(DM) & $\mathbf{47 \pm 3}$ & $\mathbf{52 \pm 3}$ & $8 \pm 2$ & $10 \pm 1$ & $\mathbf{51 \pm 2}$ & $6 \pm 1$\\
\bottomrule[1.46pt]
\end {tabular}\par
\end{sc}
\end{center}
\vskip -0.1in
\label{table:full}
\end{table*}

\begin{table*}[t]
\caption{Final performance on benchmark tasks. The setup is the same as in Table \ref{table:pg}. CV (ours) achieves consistent gains over deterministic policy $+$ \gls{ES} as well as more advanced baselines such as TRPO and PPO. We also include a \gls{PG} baseline for easy comparison. In the following, 'Det' denotes the deterministic policies.}
\vskip 0.1in
\begin{center}
\footnotesize
\begin{sc}
\begin{tabular}{ C{0.9in} *6{C{.85in}}}\toprule[1.5pt]
\bf Tasks & \bf PPO  & \bf TRPO & \bf Det &  \bf CV (Ours) & \bf PG \\\midrule
%LQR      &  $-176 \pm 12$  & $-1337 \pm 573$ & $-1246 \pm 502$ & $-5634 \pm 1059$ &  $\mathbf{-143 \pm 4}$ & $-7243 \pm 275$ \\
Swimmer      &  $11 \pm 7$  & $11 \pm 7$ & $191 \pm 100$  &  $\mathbf{237 \pm 33}$ & $-132 \pm 5$\\
HalfCheetah & $-175 \pm 20$ & $-174 \pm 16$ & $1645 \pm 1298$  & $\mathbf{1897 \pm 232}$ & $-180 \pm 4$  \\
Walker & $657 \pm 291$ & $657 \pm 292$ & $1588\pm 744$ & $\mathbf{1476 \pm 112}$ & $282 \pm 25$ \\
Pong(R) & $-17.1 \pm 0.4$ & $-15.0 \pm 2.6$ & $-10.4 \pm 5.4$  & $\mathbf{-3.0 \pm 0.3}$ & $-17 \pm 0.2$\\
HalfCheetah(R) & $14 \pm 2$ & $13 \pm 3$ & $502 \pm 199$ & $\mathbf{709 \pm 16}$ & $12 \pm 0$\\
BipedalWalker & $-66 \pm 39$ & $-66 \pm 38$ & $1 \pm 2$  & $\mathbf{105 \pm 40}$ & $-82 \pm 12$ \\
%Cheetah(DM) & $20 \pm 20$ & $36 \pm 32$ & $?$ & $\mathbf{296 \pm 15}$ & $25 \pm 6$ \\
Pendulum(DM) & $0.6 \pm 0.5$ & $7.5 \pm 4.8$ & $40 \pm 16$ & $\mathbf{43 \pm 1}$ & $3 \pm 1$ \\
%TwoPoles(DM) & $264 \pm 46$ & $42 \pm 52$ & $?$ & $\mathbf{245 \pm 29}$ & $14 \pm 1$ \\
%Swingup(DM) & $\mathbf{394 \pm 15}$ & $\mathbf{369 \pm 22}$ & $\mathbf{414 \pm 31}$ & $67 \pm 14$ & $\mathbf{406 \pm 26}$ & $55 \pm 10$ \\
Balance(DM) & $264 \pm 46$ & $515 \pm 42$ & $692 \pm 250$ & $\mathbf{847 \pm 71}$ & $401 \pm 12$\\
HopperHop(DM) & $0.2 \pm 0.2$ & $0.0 \pm 0.0$ & $4.6 \pm 6.2$  & $\mathbf{6.5 \pm 1.5}$ & $0.1 \pm 0.0$ \\
%Stand(DM) & $21 \pm 5$ & $36 \pm 10$ & $\mathbf{54 \pm 4}$ & $1.0 \pm 0.2$ & $\mathbf{60 \pm 11}$ & $0.5 \pm 0.1$\\
AntWalk(DM) & $96 \pm 36$ & $180 \pm 41$ & $192 \pm 20$  & $\mathbf{239 \pm 10}$ & $100 \pm 11$ \\
%AntEscape(DM) & $6 \pm 3$ & $10 \pm 1$ & $?$ & $\mathbf{51 \pm 2}$ & $6 \pm 1$\\
\bottomrule[1.46pt]
\end {tabular}\par
\end{sc}
\end{center}
\vskip -0.1in
\label{table:pg}
\end{table*}
 
\paragraph{Comparison with Policy Gradients.} A natural question is what happens when we update the policy just based on the \gls{PG} estimator? We show the complete comparison in \Cref{table:full}, where we find pure \gls{PG} to be mostly outperformed by the other \gls{ES} baselines. We speculate that this is because the vanilla \gls{PG} are themselves quite unstable, as commonly observed in prior works on \gls{PG} which aim to alleviate the instability by introducing bias in exchange for smaller variance \citep{schulman2015,mnih2016,schulman2017}. We provide a detailed review in the Appendix. This comparison also implies that a careful control variate scheme can extract the benefit of \gls{PG} estimators for variance reduction in \gls{ES}, instead of completely relying on \gls{PG}.

To assess the impact of our method, we also compare with trust-region based on-policy \gls{PG} algorithms: TRPO \citep{schulman2015} and PPO \citep{schulman2017}. Instead of a batched implementation as in the original paper, we adopt a fully on-line \gls{PG} updates to align with the way \gls{ES} baselines are implemented. We list detailed hyper-parameters and implementations in the Appendix C. Due to space limit, the results of selected tasks are presented in table \ref{table:pg} and we leave a more comprehensive set of results in the Appendix.

We see that the trust-region variants lead to generally more stable performance than the vanilla \gls{PG}. Nevertheless, the performance of these algorithms do not match those of the \gls{ES} baselines. There are several enhancements one can make to improve these \gls{PG} baselines, all generally relying on biasing the gradients in exchange for smaller variance, e.g. normalizing the observations and considering a biased objective \citep{baselines}. While here we consider only the 'unbiased' \gls{PG} methods, we discuss these methods in the Appendix.

\paragraph{Comparison with ES baselines.} As mentioned in Section 3, the CV is applicable only in cases where policies are stochastic. However, prior works consider mainly deterministic policy for continuous control \citep{mania2018simple,choromanski2018structured}. To assess the effects of policy choices, we adopt the same \gls{ES} baseline pipeline but with deterministic policies. The results of deterministic policies are reported in Table \ref{table:pg}. We make several observations: (1) Deterministic policies generally perform better than stochastic policies when using \gls{ES} baselines. We speculate this is because the additional noise introduced by the stochastic policy is not worth the additional variance in the gradient estimation. Nevertheless, CV can leverage the variance reduction effects thanks to the stochastic policy, and maintain generally the best performance across tasks. Such comparison illustrates that it is beneficial to combine stochastic policies with \gls{ES} methods as long as there is proper variance reduction.

\paragraph{Practical Scalability.} In practical applications, the scalability of algorithms over large computational architecture is critical. Since CV blends ideas from \gls{ES} with \gls{PG}, we require both distributed sample collection  \citep{salimans2017evolution,mania2018simple} and distributed gradient computation \citep{baselines}. Since both components can be optimally implemented over large architectures, we expect CV to properly scale. A more detailed discussion is in the Appendix.

\section{Conclusion}
We constructed a control variate for \gls{ES} that take advantage of the \gls{MDP} structure of \gls{RL} problems to improve on state-of-the-art 
variance reduction methods for \gls{ES} algorithms.
Training algorithms using our control variate outperform those applying general-purpose \gls{MC} methods for variance reduction. 
We provided theoretical insight into the effectiveness of our algorithm
as well as exhaustive comparison of its performance with other methods on the set of  over 16 \gls{RL} benchmark tasks. In principle, this control variate can be combined with other variance reduction techniques; this may lead to further performance gains.

We leave as future work to study how similar structured control variates can be applied to improve the performance of state-of-the-art \gls{PG} algorithms, in particular, cases where gradients have already been deliberately biased to achieve better performance such as in \citep{baselines}.

%\subsection{Discrete Domain}
%\paragraph{Benchmark Environments.} Discrete action tasks are selected from OpenAI gym \citep{brockman2016}. Certain discrete environments are derived from a continuous action task by discretizing each dimension of the $K$ action dimensions into $5$ atoms. The resulting action space has $K^5$ discrete actions.
\newpage
\bibliographystyle{apa}
\bibliography{your_bib_file.bib}

\newpage
\appendix
\onecolumn
\section{How Much Variance Reduction is Possible?}
Recall in the main paper we consider a one-step MDP with action $a\in \mathbb{R}^d$. The reward function is $\alpha^Ta$ for some $\alpha \in \mathbb{R}^d$. Consider a Gaussian policy with mean parameter $\mu$ and fixed covariance matrix $\Sigma = \sigma_2^2 \mathbb{I}$. The action is sampled as $a \sim \mathcal{N}(\mu,\Sigma)$. ES convolves the reward objective with a Gaussian with covariance matrix $\sigma_1^2 \mathbb{I}$. Let $\epsilon_1^{(i)},\epsilon_2^{(i)}\sim \mathcal{N}(0,\mathbb{I}),1\leq i\leq N$ be $N$ independent reparameterized noise, we can derive the vanilla ES estimator
\begin{align}
\hat{g}_{\mu}^{\textsc{es}} = \frac{1}{N} \sum_{i=1}^N \hat{g}_{\mu,i}^{\textsc{es}} = \frac{1}{N} \sum_{i=1}^N \alpha^T (\mu + \sigma_1 \epsilon_1^{(i)} + \sigma_2 \epsilon_2^{(i)} )\frac{\epsilon_2^{(i)}}{\sigma_2}\nonumber
\end{align}
The orthogonal estimator is constructed by $N$ perturbations $\epsilon_{2,\text{ort}}^{(i)}$ such that $\langle \epsilon_{2,\text{ort}}^{(i)},\epsilon_{2,\text{ort}}^{(j)} \rangle = 0$ for $i\neq j$, and each $\epsilon_2^{(i)}$ is still marginally $d$-dimensional Gaussian. The orthogonal estimator is
\begin{align}
\hat{g}_{\mu}^{\text{ort}} = \frac{1}{N} \sum_{i=1}^N \hat{g}_{\mu,i}^{\text{ort}} = \frac{1}{N} \sum_{i=1}^N \alpha^T (\mu + \sigma_1 \epsilon_1^{(i)} + \sigma_2 \epsilon_{2,\text{ort}}^{(i)} )\frac{\epsilon_{2,\text{ort}}^{(i)}}{\sigma_2}\nonumber
\end{align}
Finally, we consider the ES gradient estimator with control variate. In particular, we have the reparameterized gradient as
\begin{align}
    \hat{g}_{\mu}^{\textsc{re}} = \frac{1}{N} \sum_{i=1}^N \hat{g}_{\mu,i}^{\textsc{re}} = \frac{1}{N} \sum_{i=1}^N \alpha^T (\mu + \sigma_1 \epsilon_1^{(i)} + \sigma_2 \epsilon_2^{(i)} )\frac{\epsilon_1^{(i)}}{\sigma_1}\nonumber
\end{align}
The general gradient estimator with control variate is
\begin{align}
\hat{g}_{\mu}^{\textsc{cv}} = \hat{g}_{\mu}^{\textsc{es}} + \eta \odot (\hat{g}_{\mu}^{\textsc{re}} - \hat{g}_{\mu}^{\textsc{es}})\nonumber
\end{align}
where $\eta \in \mathbb{R}^d$. Since $\eta$ can be indepdently chosen across dimensions, the maximal variance reduction is achieved by setting $\eta_i =- \frac{\text{cov}(X_i,Y_i)}{\mathbb{V}[Y_i]}$ where here $X = \hat{g}_{\mu}^{\textsc{es}}, Y = \hat{g}_{\mu}^{\textsc{re}} - \hat{g}_{\mu}^{\textsc{es}}$.

Recall that for a vector $g$ of dimension $d$, its variance is defined as the sum of the variance of its components $\mathbb{V}[g] = \sum_{i=1}^d \mathbb{V}[g_i]$.
For simplicity, let $\rho = \frac{\sigma_2}{\sigma_1}$.
We derive the variance for each estimator below.

\paragraph{Vanilla ES.} For the vanilla ES gradient estimator, the variance is
\begin{align}
   \mathbb{V}[ \hat{g}_{\mu}^{\textsc{es}}  ]  = \frac{d+1}{N} \Vert \alpha \Vert_2^2\nonumber
\end{align}
\paragraph{Orthogonal ES.} For the orthogonal ES gradient estimator, the variance is
\begin{align}
   \mathbb{V}[ \hat{g}_{\mu}^{\text{ort}}]  = \frac{(1+\rho^2)d+2-N}{N} \Vert \alpha \Vert_2^2\nonumber
\end{align}

\paragraph{ES with Control Variate.} For the ES gradient estimator with control variate, recall the above notation $X = \hat{g}_{\mu}^{\textsc{es}}, Y = \hat{g}_{\mu}^{\textsc{re}} - \hat{g}_{\mu}^{\textsc{es}}$. We first compute $\rho(X_p,Y_p)^2 = \frac{\text{cov}^2(X_p,Y_p)}{\mathbb{V}[X_p] \mathbb{V}[Y_p]}$ for each component $p$. Let $X_{p,i},Y_{p,i}$ be the $p$th component of $ \hat{g}_{\mu,i}^{\textsc{es}} $ and $ \hat{g}_{\mu,i}^{\textsc{re}}-\hat{g}_{\mu,i}^{\textsc{es}} $ respectively. We will detail how to compute $\text{cov}(X_p, Y_p),\mathbb{V}[V_p]$ in the next section. With these components in hand, we have the final variance upper bound
\begin{align}
    \mathbb{V}[ \hat{g}_{\mu}^{\textsc{cv}}] &\leq \mathbb{V}[ \hat{g}_{\mu}^{\textsc{es}}]  \{1 - \frac{(1+\rho^2)[d((1+\rho^2) - 4]}{[(1+\rho^2)d + 1](2 + \rho^2 + \frac{1}{\rho^2})} \}.\nonumber
\end{align}

\section{Derivation Details}
Recall that for a vector $g$ of dimensioon $d$, we define its variance as $\mathbb{V}[g] = \sum_{i=1}^d \mathbb{V}[g_i]$
For simplicity, recall that $\rho = \frac{\sigma_2}{\sigma_1}$.
\subsection{Variance of Orthogonal ES}
We derive the variance of orthogonal ES based on the formula in the Appendix of \citep{choromanski2018structured}. In particular, we can easily compute the $i$ sample estimate for the $p$th component of $X_{i,p} = [\hat{g}_{\mu,i}^{\text{ort}}]_p$
\begin{align}
    \mathbb{E}[X_{i,p}^2] = (1+\rho^2) \Vert \alpha\Vert_2^2 + \alpha_p^2 \nonumber
\end{align}
Hence the variance can be calculated as
\begin{align}
    \mathbb{V}[\hat{g}_{\mu}^\text{ort}] = \mathbb{V}[X] = \frac{(1+\rho^2) d + 2 - N}{N} \Vert \alpha\Vert_2^2\nonumber
\end{align}

\subsection{Variance of Vanilla ES}
When we account for the cross product terms as in \citep{choromanski2018structured}, we can easily derive
\begin{align}
    \mathbb{V}[\hat{g}_{\mu}^\textsc{es}] = \mathbb{V}[X] = \frac{(1+\rho^2) d + 1}{N} \Vert \alpha\Vert_2^2.\nonumber
\end{align}
We can also easily derive the variance per component $\mathbb{V}[X_p] = \frac{1}{N}((1+\rho)^2 \Vert \alpha\Vert_2^2 + \alpha_p^2)$.

\subsection{Variance of ES with Control Variate}
Recall the definition $X_p = X^T e_p, Y_p = Y^T e_p$ where $e_p$ is a one-hot vector with $[e_p]_i = \delta_{ip}$. For simplicity, we fix $p$ and denote $x_i = X_{p,i}, y_i = X_{p,i} - Y_{p,i}$.
\paragraph{Step 1: Calculate $\text{cov}(X_p,Y_p)$.} The notation produces the covariance
\begin{align}
    \text{cov}(X_p,Y_p) &= \text{cov}(\frac{1}{N}\sum_{i=1}^N x_i, \frac{1}{N}\sum_{i=1}^N (x_i - y_i)) \nonumber \\
    &= \frac{1}{N^2} \mathbb{E}[\sum_{i,j} x_ix_j - x_iy_j].  \nonumber \\
\end{align}
We identify some necessary components. Let $i\neq j$, then
\begin{align}
    \mathbb{E}[x_i^2]
    &= \mathbb{E}[(\alpha^T (\sigma_1 \epsilon_1 + \sigma_2 \epsilon_2)\frac{\epsilon_{1,p}}{\sigma_1})^2]   \nonumber \\
    &= \mathbb{E}[(\alpha^T \epsilon_1)^2 \epsilon_{1,p}^2 + (\alpha^T \epsilon_2)^2 \rho^2]   \nonumber \\
    &= (1+\rho^2) \Vert \alpha \Vert_2^2 + 2\alpha_p^2   \nonumber \\
      \mathbb{E}[x_ix_j] &= \mathbb{E}[x_iy_j] = \alpha_p^2  \nonumber \\
      \mathbb{E}[x_iy_i] &= \mathbb{E}[(\alpha^T (\sigma_1\epsilon_1 + \sigma_2\epsilon_2)^2 \frac{\epsilon_{1,p}\epsilon_{2,p}}{\sigma_1\sigma_2}] \nonumber \\
    &= \mathbb{E}[2\alpha^T \epsilon_1 \alpha^T \epsilon_2 \epsilon_{1,p}\epsilon_{2,p}]
    = 2\alpha_p^2 \nonumber \\
\end{align}
We can hence derive
\begin{align}
    \text{cov}(X_p,Y_p) &= \frac{1}{N^2} [\sum_{i=1}^N \mathbb{E}[ x_i^2 - x_iy_i] + \sum_{i\neq j} \mathbb{E}[x_ix_j - x_iy_j]] \nonumber \\
  &= \frac{1}{N} [(1+\rho^2) \Vert \alpha \Vert_2^2 - 2\alpha_p^2]
 \nonumber
\end{align}
\paragraph{Step 2: Calculate $\mathbb{V}[Y_p]$.} We only need to derive $\mathbb{E}[Y_{p,i}^2] = \mathbb{E}[(\alpha^T(\sigma_1 \epsilon_1  + \sigma_2 \epsilon_2) (\frac{\epsilon_{1,p}}{\sigma_1} - \frac{\epsilon_{2,p}}{\sigma_2})^2]$. After expanding all the terms, we can calculate
\begin{align}
    \mathbb{E}[(\alpha^T(\sigma_1 \epsilon_1  + \sigma_2 \epsilon_2) (\frac{\epsilon_{1,p}}{\sigma_1} - \frac{\epsilon_{2,p}}{\sigma_2})^2] &= (2 + \rho^2 + \frac{1}{\rho^2}) \Vert \alpha \Vert_2^2 \nonumber
\end{align}

\paragraph{Step 3: Combine all components.} We finally combine all the previous elements into the main result on variance reduction. Assuming that the scaling factor of the control variate $\eta$ is optimally set, the maximum variance reduction leads to the following resulting variance of component $p$. Using the above notations
\begin{align}
    \mathbb{V}[[\hat{g}_{\mu}^{\textsc{cv}}]_p] &= \mathbb{V}[X_p] - \frac{\text{cov}^2(X_p, Y_p)}{\mathbb{V}[Y_p]} \nonumber \\
    &= \frac{(1+\rho)^2\Vert \alpha\Vert_2^2 + \alpha_p^2}{N} - \frac{1}{N} \frac{[(1+\rho)^2 \Vert\alpha\Vert_2^2 -2\alpha_p^2]^2}{(2+\rho^2+\frac{1}{\rho^2}) \Vert\alpha\Vert_2^2}. \nonumber
\end{align}
We can lower bound the right hand side and sum over $d$ dimensions,
\begin{align}
    \mathbb{V}[\hat{g}_{\mu}^{\textsc{cv}}] &= \sum_{p=1}^d  \mathbb{V}[[\hat{g}_{\mu}^{\textsc{cv}}]_p] \leq \mathbb{V}[X] - \nonumber \\  &\frac{d}{N}\frac{(1+\rho^2)^2}{2+\rho^2+\frac{1}{\rho^2}} \Vert \alpha\Vert_2^2 + \frac{4}{N}\frac{1+\rho^2}{2+\rho^2+\frac{1}{\rho^2}} \nonumber
\end{align}
Finally, we plug in $\mathbb{V}[X]$ and calculate the variance ratio with respect to the vanilla ES
\begin{align}
    \frac{\mathbb{V}[\hat{g}_\mu^{\textsc{cv}}]}{\mathbb{V}[\hat{g}_\mu^{\textsc{es}}]} \leq 1 - \frac{\rho^2[d((1+\rho^2) - 4]}{[(1+\rho^2)d + 1](1+\rho^2)}.  \nonumber
\end{align}
As a comparison, we can calculate the variance ratio of the orthogonal ES
\begin{align}
    \frac{\mathbb{V}[\hat{g}_\mu^{\text{ort}}]}{\mathbb{V}[\hat{g}_\mu^{\textsc{es}}]} = \frac{(1+\rho^2)d +2 - N}{(1+\rho^2)d + 1}.  \nonumber
\end{align}

\paragraph{When does the control variate achieve lower variance?}
We set the inequality $\mathbb{V}[\hat{g}_\mu^{\text{ort}}] \geq \mathbb{V}[\hat{g}_\mu^{\textsc{cv}}]$ and calculate the following condition
\begin{align}
    \rho \geq \rho_0 = \sqrt{\frac{N+3-d + \sqrt{(d-N-3)^2 + 4(N-1)d}}{2d}}.
    \label{eq:rho0}
\end{align}
The expression (\ref{eq:rho0}) looks formidable. To simplify the expression, consider the limit $N\rightarrow \infty$ while maintaining $\frac{N}{d} \in [0,1]$. Taking this limit allows us to drop certain constant terms on the right hand side, which produces $\rho_0 = \sqrt{\frac{N}{d}}$.

\section{Additional Experiment Details}
\subsection{Updating Discount Factor $\gamma$}  The discount factor $\gamma$ is updated using \gls{ES} methods. Specifically, at each iteration $t$, we maintain a current $\gamma_t$. The aim is to update $\gamma_t$ such that the empirical estimate of $\mathbb{V}[\hat{g}_\theta^{\textsc{cv}}]$ is minimized. For each value of $\gamma$ we can calculate a $\hat{g}_\theta^{\textsc{cv}}(\gamma)$, here we explicitly note its dependency on $\gamma$. For this iteration, we setup a blackbox function as $F(\gamma) = \sum_{i=1}^d [\hat{g}_\theta^{\textsc{cv}}(\gamma)]_i^2$ where $d$ is the dimension of parameter $\theta$. Then the \gls{ES} update for $\gamma$ is $\gamma_{t+1} = \gamma_t - \alpha_\gamma \hat{g}_\gamma $, where
\begin{align}
    \hat{g}_\gamma = \frac{1}{N_\gamma} \sum_{i=1}^{N_\gamma} \frac{F(\gamma_t + \sigma_\gamma \epsilon_i)}{\sigma_\gamma} \epsilon_i, \ \epsilon_i \sim \mathcal{N}(0,1).
\end{align}
As mentioned in the paper, to ensure $\gamma \in (0,1]$ we parameterize $\gamma = 1 - \exp(\phi)$ in our implementation. We optimize $\phi$ using the same \gls{ES} scheme but in practice we set $\alpha_\phi,\sigma_\phi,N_\phi$. Here we have $\alpha_\phi \in \{10^{-4},10^{-5},10^{-6}\}$, $\sigma_\phi = 0.02$ and $N_\phi = 10$. Sometimes we find it effective to just set $\gamma$ to be the initial constant, because the control variate scalar $\eta$ is also adjusted to minimize the variance.

\subsection{Normalizing the gradients} Vanilla stochastic gradient updates are sensitive to the scaling of the objective function, which in our case are the reward functions. Recall the vanilla \gls{ES} estimator
$
    \hat{g}_\theta^{\text{es}} = \frac{1}{N} \sum_{i=1}^N \frac{J(\pi_{\theta+\sigma\cdot\epsilon_i})}{\sigma} \epsilon_i
$
where $J(\pi_\theta)$ is the return estimate under policy $\pi_\theta$. To ensure that the gradent is properly normalized, the common practice \citep{salimans2017evolution,mania2018simple,choromanski2018structured} is to normalize the returns $\tilde{J} \leftarrow\frac{J -\bar{J}}{\sigma(J)}$, where $\bar{J},\sigma(J)$ are the mean/std of the estimated returns of the batch of $N$ samples. The normalized returns $\tilde{J}$ are used in place of the original returns in calculating the gradient estimators. In fact, we can interpret the normalization as subtracting a moving average baseline (for variance reduction) and dividing by a rough estimate of the local Lipchitz constant of the objective landscape. These techniques tend to make the algorithms more stable under reward functions with very different scales.

The same technique can be applied to the control variate. We divide the original control variate by $\sigma(J)$ to properly scale the estimators.

\paragraph{Details on policy gradient methods.}
\gls{PG} methods are implemented following the general algorithmic procedure as follows: collect data using a previous policy $\pi_{\theta_{t-1}}$, construct loss function and update the policy parameter using gradient descent $\theta_{t} \leftarrow $ to arrive at $\pi_{\theta_t}$, then iterate. We consider three baselines: vanilla \gls{PG}, TRPO and PPO. In our implementation, these three algorithms differ in the construction of the loss function. For vanilla \gls{PG}, the loss function is $L = -\mathbb{E}_{s,a}[{\frac{\pi_\theta(a|s)}{\pi_{\theta_\text{old}}}(a|s)} A(s,a)]$ where $A(s,a)$ is the advantage estimtion and $\theta_{\text{old}}$ is the prior policy iterate. For PPO, the loss function is $L = -\mathbb{E}_{s,a}[{\text{clip}\{\frac{\pi_\theta(a|s)}{\pi_{\theta_\text{old}}}(a|s)},1-\epsilon,1+\epsilon\} A(s,a)]$ where $\text{clip}\{x,a,b\}$ is to clip $x$ between $a$ and $b$ and $\epsilon=0.2$. In practice, we go one step further and implement the action dependent clipping strategy as in \citep{schulman2017}. For TRPO, we implement the dual optimization objective of the original formulation \citep{schulman2015} and set the loss function $L = -\mathbb{E}_{s,a}[{\frac{\pi_\theta(a|s)}{\pi_{\theta_\text{old}}}(a|s)} A(s,a)] + \eta \mathbb{E}_s\big[\mathbb{KL}[\pi_\theta(\cdot|s)|| \pi_{\theta_{\text{old}}}(\cdot|s)]\big]$ where we select $\eta \in \{0.1,1.0\}$. 

When collecting data, we collect $2N$ full episodic trajectories as with the \gls{ES} baselines. This is different from a typical implementation \citep{baselines}, where \gls{PG} algorithms collect a fixed number of samples per iteration. Also, we take only one gradient descent on the surrogate objective per iteration, as opposed to multiple updates. This reduces the policy optimization procedure to a fully on-line fashion as with the \gls{ES} methods.

\section{Additional Experiments}

\subsection{Additional Baseline Comparison}

\begin{figure*}[h]
\centering
\subfigure[\textbf{Balance (DM)}]{\includegraphics[width=.23\linewidth]{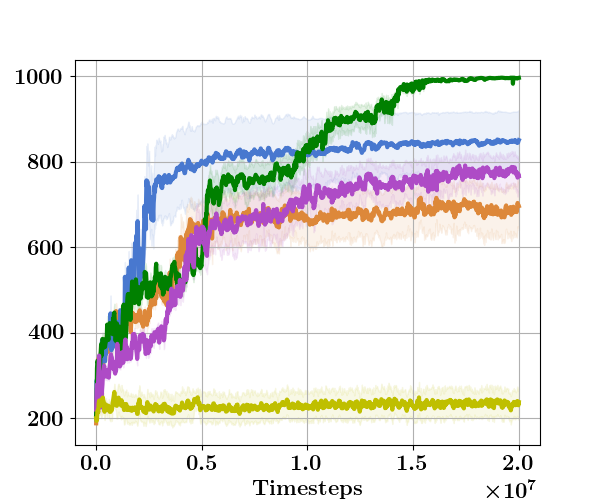}}
\subfigure[\textbf{Pendulum (DM)}]{\includegraphics[width=.23\linewidth]{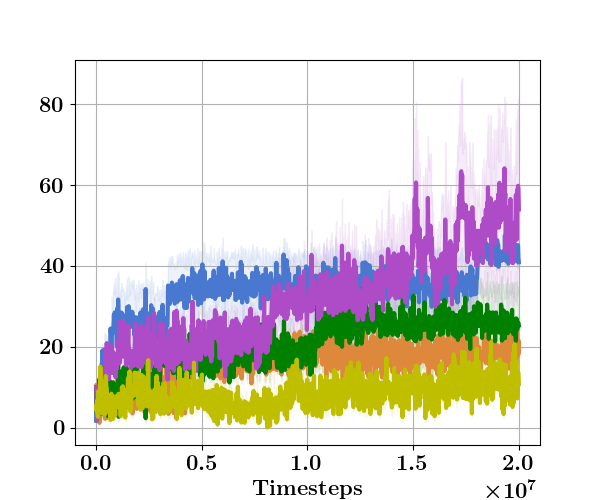}}
%\subfigure[\textbf{CheetahRun (DM)}]{\includegraphics[width=.23\linewidth]{plots/esrank_long_dmcheetahrun.png}}
%\subfigure[\textbf{HopperStand (DM)}]{\includegraphics[width=.23\linewidth]{plots/es_long_dmhopperstand.png}}
\subfigure[\textbf{HopperHop (DM)}]{\includegraphics[width=.23\linewidth]{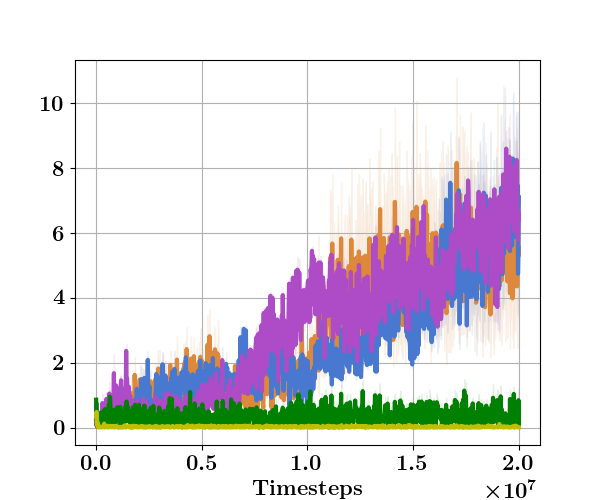}}
\subfigure[\textbf{AntWalk(DM)}]{\includegraphics[width=.23\linewidth]{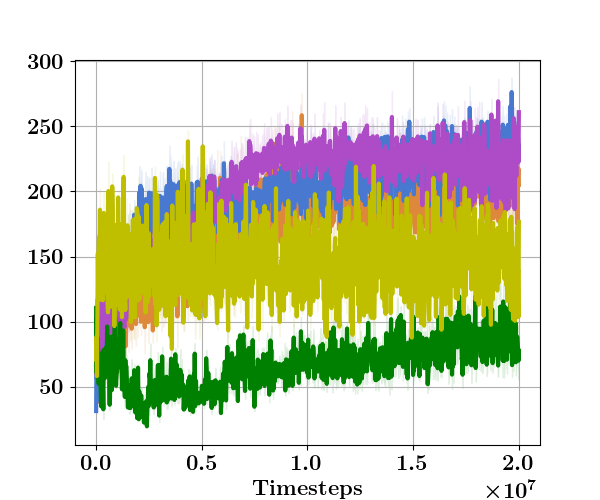}}
%\subfigure[\textbf{AntEscape(DM)}]{\includegraphics[width=.23\linewidth]{plots/esrank_long_dmantescape.png}}
\caption{\small{Training performance on Continuous Control Benchmarks: Swimmer, HalfCheetah, CartPole $+$ $\{\text{Swingup},\text{TwoPoles},\text{Balance}\}$, Pendulum Swingup, Cheetah Run and Hopper. Tasks with DM stickers are from the DeepMind Control Suites. We compare five alternatives: baseline ES (orange), CV (blue, ours), ORTHO (marron), GCMC (green) and QMC (yellow). Each task is trained for $2 \cdot 10^7$ time steps (LQR is trained for $4 \cdot 10^6$ steps) and the training curves show the mean $\pm$ std cumulative rewards across 5 random seeds.}}
\label{figure:objective2}
\end{figure*}

Due to space limit, we omit the baseline comparison result on four control tasks from the main paper. We show their results in Figure \ref{figure:objective2}. The setup is exactly the same as in the main paper.

\subsection{Discussion on Policy Gradient Methods}

\paragraph{Instability of \gls{PG}.} It has been observed that the vanilla \gls{PG} estimators are not stable. Even when the discount factor $\gamma < 1$ is introduced to reduce the variance, vanilla \gls{PG} estimators can still have large variance due to the long horizon. As a result in practice, the original form of the \gls{PG} (\ref{eq:pg}) is rarely used. Instead, prior works and practical implementations tend to introduce bias into the estimators in exchange for lower variance: e.g. average across states intsead of trajectories \citep{baselines}, clipping based trust region \citep{schulman2017} and biased advantage estimation \citep{schulman2015high}. These techniques stabilize the estimator and lead to state-of-the-art performance, however, their theoretical property is less clear (due to their bias). We leave the study of combining such biased estimators with \gls{ES} as future work.

\subsection{Additional Results with TRPO/PPO and Deterministic Policies}
We provide additional comparison results against TRPO/PPO and deterministic policies in Table \ref{table:pg2}. We see that generally TRPO/PPO achieve better performance, yet are still under-performed by CV. On the other hand, though deterministic policies tend to outperform stochastic policies when combined with the \gls{ES} baselines, they are not as good as stochastic policies with CV.

\paragraph{Comparison of implementation variations of \gls{PG} algorithms.} Note that for fair comparison, we remove certain functionalities of a full fledged \gls{PG} implementation as in \citep{baselines}. We identify important differences between our baseline and the full fledged implementation and evaluate their effects on the performance in Table \ref{table:pg3}. Our analysis focuses on PPO. These important implementation differences are
\begin{itemize}
    \item \textbf{S}: Normalization of observations.
    \item \textbf{A}: Normalization of advantages.
    \item \textbf{M}: Multiple gradient updates (in particular, $10$ gradient updates), instead of $1$ for our baseline.
\end{itemize}
To understand results in Table \ref{table:pg3}, we start with the  notations of variations of PPO implementations. The PPO denotes our baseline for comparison; the PPO+S denotes our baseline with observation normalization. Similarly, we have PPO+A and PPO+M. Notations such as PPO+S+A denote the composition of these implementation techniques. We evaluate these implementation variations on a subset of tasks and compare their performance in Table \ref{table:pg3}.

We make several observations: \textbf{(1)} Multiple updates bring the most significant for \gls{PG}. While conventional \gls{ES} implementations only consider one gradient update per iteration \citep{salimans2017evolution}, it is likely that they coudl also benefit from this modification; \textbf{(2)} Even when considering these improvements, our proposed CV method still outperforms \gls{PG} baselines on a number of tasks.

\section{Discussion on Scalability}

 \gls{ES} easily scale to a large distributed architecture for the trajectory collection \citep{salimans2017evolution,mania2018simple}. Indeed, the bottleneck of most \gls{ES} applications seem to be the sample collection, which can be conveniently distributed across multiple cores (machines). Fortunately, many open source implementations have provided high-quality packages via efficient inter-process communications and various techniques to reduce overheads \citep{salimans2017evolution,mania2018simple}.
 
 On the other hand, distributing \gls{PG} requires more complicated software design. Indeed, distributed \gls{PG} involve both distributed sample collection and gradient computations. The challenge lies in how to construct gradients via multiple processes and combine them into a single update, without introducing costly overheads. Open source implementations such as \citep{baselines} have achieved synchronized distributed gradient computations via MPI.
 
 To scale CV, we need to combine both distributed sample collection from \gls{ES} and distributed gradient computation from \gls{PG}. Since both of the above two components have been implemented with high quality via open source packages, we expect it not to be a big issue to scale CV, in particular to combine scalable gradient computation with \gls{ES}. However, this requires additional efforts - as these two parts have never been organically combined before and there is so far little (open source) engineering effort into this domain. We leave this as future work.
 
\begin{table*}[t]
\caption{Final performance on benchmark tasks. The policy is trained for a fixed number of steps on each task. The result is $\text{mean} \pm \text{std}$ across 5 random seeds. The best results are highlighted in bold font. We highlight multiple methods if their results cannot be separated ($\text{mean} \pm \text{std}$ overlap). CV (ours) achieves consistent gains over the baseline and other variance reduction methods. We also include a \gls{PG} baseline.}
\vskip 0.1in
\begin{center}
\footnotesize
\begin{sc}
\begin{tabular}{ C{0.9in} *6{C{.7in}}}\toprule[1.5pt]
\bf Tasks & \bf PPO  & \bf TRPO & \bf Det &  \bf CV (Ours) & \bf PG \\\midrule
%LQR      &  $-176 \pm 12$  & $-1337 \pm 573$ & $-1246 \pm 502$ & $-5634 \pm 1059$ &  $\mathbf{-143 \pm 4}$ & $-7243 \pm 275$ \\
Swimmer      &  $11 \pm 7$  & $11 \pm 7$ & $191 \pm 100$  &  $\mathbf{237 \pm 33}$ & $-132 \pm 5$\\
HalfCheetah & $-175 \pm 20$ & $-174 \pm 16$ & $1645 \pm 1298$  & $\mathbf{1897 \pm 232}$ & $-180 \pm 4$  \\
Walker & $657 \pm 291$ & $657 \pm 292$ & $1588\pm 744$ & $\mathbf{1476 \pm 112}$ & $282 \pm 25$ \\
Pong(R) & $-17.1 \pm 0.4$ & $-15.0 \pm 2.6$ & $-10.4 \pm 5.4$  & $\mathbf{-3.0 \pm 0.3}$ & $-17 \pm 0.2$\\
HalfCheetah(R) & $14 \pm 2$ & $13 \pm 3$ & $502 \pm 199$ & $\mathbf{709 \pm 16}$ & $12 \pm 0$\\
BipedalWalker & $-66 \pm 39$ & $-66 \pm 38$ & $1 \pm 2$  & $\mathbf{105 \pm 40}$ & $-82 \pm 12$ \\
Cheetah(DM) & $20 \pm 20$ & $36 \pm 32$ & $241 \pm 47$ & $\mathbf{296 \pm 15}$ & $25 \pm 6$ \\
Pendulum(DM) & $0.6 \pm 0.5$ & $7.5 \pm 4.8$ & $40 \pm 16$ & $\mathbf{43 \pm 1}$ & $3 \pm 1$ \\
TwoPoles(DM) & $264 \pm 46$ & $42 \pm 52$ & $190 \pm 58$ & $\mathbf{245 \pm 29}$ & $14 \pm 1$ \\
%Swingup(DM) & $\mathbf{394 \pm 15}$ & $\mathbf{369 \pm 22}$ & $\mathbf{414 \pm 31}$ & $67 \pm 14$ & $\mathbf{406 \pm 26}$ & $55 \pm 10$ \\
Balance(DM) & $264 \pm 46$ & $515 \pm 42$ & $692 \pm 250$ & $\mathbf{847 \pm 71}$ & $401 \pm 12$\\
HopperHop(DM) & $0.2 \pm 0.2$ & $0.0 \pm 0.0$ & $4.6 \pm 6.2$  & $\mathbf{6.5 \pm 1.5}$ & $0.1 \pm 0.0$ \\
%Stand(DM) & $21 \pm 5$ & $36 \pm 10$ & $\mathbf{54 \pm 4}$ & $1.0 \pm 0.2$ & $\mathbf{60 \pm 11}$ & $0.5 \pm 0.1$\\
AntWalk(DM) & $96 \pm 36$ & $180 \pm 41$ & $192 \pm 20$  & $\mathbf{239 \pm 10}$ & $100 \pm 11$ \\
AntEscape(DM) & $6 \pm 3$ & $10 \pm 1$ & $13 \pm 8$ & $\mathbf{51 \pm 2}$ & $6 \pm 1$\\
\bottomrule[1.46pt]
\end {tabular}\par
\end{sc}
\end{center}
\vskip -0.1in
\label{table:pg2}
\end{table*}

\begin{table*}[t]
\caption{Evaluation of implementation variations of \gls{PG} methods, with comparison to the CV. Below show the final performances on benchmark tasks. The policy is trained for a fixed number of steps on each task. The result below shows the mean performance averaged across 3 random seeds. The best results are highlighted in bold font. Notice below the table is transposed due to space.}
\vskip 0.1in
\begin{center}
\footnotesize
\begin{sc}
\begin{tabular}{ C{0.9in} *6{C{.7in}}}\toprule[1.5pt]
 \bf Tasks &  Swimmer  &  HalfCheetah &  Pendulum &   Balance &  Hopper &  AntWalk  \\\midrule
%LQR      &  $-176 \pm 12$  & $-1337 \pm 573$ & $-1246 \pm 502$ & $-5634 \pm 1059$ &  $\mathbf{-143 \pm 4}$ & $-7243 \pm 275$ \\
\bf PPO         &         $ 11$         &     $-175$        & $0.6$ &                   $264$             & $0.2    $&$     96$ \\
\bf PPO+S        & $      30             $&$  126$&$          2                $&$       308$&$              0.5    $&$      80 $ \\
\bf PPO+A            &$  33 $&$              67$&$            26             $&$        278$&$              0.4     $&$     117 $\\
\bf PPO+S+A   &$       31$&$              192       $&$   1   $&$                     386   $&$           0.6        $&$ 106 $\\              
\bf PPO+S+M   & $      110           $&$  1579 $&$       36                 $&$     652     $&$         0.3       $&$  180 $\\ 
\bf PPO+A+M      &$   51$&$              \mathbf{2414} $&$      \mathbf{223}                 $&$ 771 $&$             46         $&$ 173 $\\      
\bf PPO+S+A+M   &$ 106$&$             1528       $&$  17$&$                     750   $&$       \mathbf{46.4}     $&$ 144 $\\ 
\bf CV(Ours)       &$    \mathbf{237}           $&$ 1897   $&$     43         $&$           \mathbf{847}       $&$     6.5         $&$ \mathbf{239} $\\
\bottomrule[1.46pt]
\end {tabular}\par
\end{sc}
\end{center}
\vskip -0.1in
\label{table:pg3}
\end{table*}

\end{document}